\pdfoutput=1

\documentclass[11pt]{article}

\usepackage[preprint]{acl}

\usepackage{times}
\usepackage{latexsym}

\usepackage[T1]{fontenc}

\usepackage[utf8]{inputenc}

\usepackage{microtype}

\usepackage{inconsolata}

\usepackage{graphicx}

\usepackage{amsmath}
\usepackage{amssymb}
\usepackage{multirow}
\usepackage{xcolor}
\usepackage{framed}
\usepackage{subcaption}
\usepackage{booktabs}

\usepackage{listings}
\usepackage{newfloat}
\DeclareFloatingEnvironment[%
    listname={List of Prompts},
    name=Prompt,
    placement=tbh,
    within=section
]{prompt}

\usepackage{textcomp}

\usepackage{xspace}

\newcommand{\vsj}{$\mathrm{VS}_{\mathrm{jina}}$\xspace}
\newcommand{\vsn}{$\mathrm{VS}_{n\textrm{-gram}}$\xspace}
\newcommand{\te}{$\mathrm{TE}_{\mathrm{jina}}$\xspace}

\makeatletter
\ifacl@finalcopy\fi
\makeatother

\title{Mind the Gap: Conformative Decoding to Improve Output Diversity of Instruction-Tuned Large Language Models}

\author{%
  Max Peeperkorn \\
  School of Computing \\
  University of Kent \\
  \texttt{m.peeperkorn@kent.ac.uk} \\
  \And
  Tom Kouwenhoven \\
  Leiden Institute of Advanced Computer Science \\
  Universiteit Leiden \\
  \texttt{t.kouwenhoven@liacs.leidenuniv.nl} \\
  \AND
  Dan Brown \\
  Cheriton School of Computer Science \\
  University of Waterloo \\
  \texttt{dan.brown@waterloo.ca} \\
  \And
  Anna Jordanous \\
  School of Computing \\
  University of Kent \\
  \texttt{a.k.jordanous@kent.ac.uk}
}

\begin{document}
\maketitle

\begin{abstract}
Instruction-tuning large language models (LLMs) reduces the diversity of their outputs, which has implications for many tasks, particularly for creative tasks.
This paper investigates the ``diversity gap'' for a writing prompt narrative generation task.
This gap emerges as measured by current diversity metrics for various open-weight and open-source LLMs. 
The results show significant decreases in diversity due to instruction-tuning.
We explore the diversity loss at each fine-tuning stage for the OLMo and OLMo 2 models to further understand how output diversity is affected.
The results indicate that DPO has the most substantial impact on diversity.
Motivated by these findings, we present a new decoding strategy, conformative decoding, which guides an instruct model using its more diverse base model to reintroduce output diversity.
We show that conformative decoding typically increases diversity and even maintains or improves quality.
\end{abstract}

\section{Introduction}

Instruction-tuned large language models (LLMs) have been shown to produce less diverse outputs than their base models \citep{hamalainen-tavast-kunnari-2023,le-bronnec-etal-2024,kirk-etal-2024,chu-etal-2024}.
The instruct model may have a bias towards certain answers or topics:
For example, prompting an LLM to write about an ``evil empire'' may disproportionally generate outputs related to Star Wars, but following instructions and engaging in conversation is an extremely useful application of LLM.
Conversely, the base model can only rely on pre-existing token patterns in the training data to generate an output of the desired artefact class.
Few-shot prompting \citep{brown-etal-2020} can push the base model to more consistent behaviour but requires a specific (diverse) set of examples.
However, such a set of the target artefact class may not be available, or the user may not know what they are looking for.

Output diversity is a desired characteristic of LLMs and other generative models \citep{tevet-berant-2021}.
This is not limited to mundane interactions, such as ``Any tips for things to do in Edinburgh?'', but also for more complex or creative tasks, such as narrative generation.
It is valuable for an LLM to respond in different ways, not only to avoid repetitiveness in conversations but also to enable (creative) exploration.
Diversity (as \textit{variety, divergence, and experimentation}) is considered a key component of creativity identified by \citet{jordanous-keller-2016} in an analysis of relevant academic literature spanning over sixty years.
Having a diverse set of options enables consideration of multiple perspectives before selecting the most suitable output \citep{grace-maher-2016,ibarrola-grace-2024}.
Experiments also suggest that humans collaboratively writing with LLMs leads to reduced diversity in the produced content \citep{padmakumar-he-2024}.
The loss of diversity due to instruction-tuning might have implications in applications such as agent-based simulations \citep[e.g.,][]{park-etal-2023}, employing the LLM-as-a-judge \citep[e.g.,][]{goes-etal-2022}, or when inducing personalities \citep{jiang-etal-2023}, but also for generating synthetic instruction-tuning datasets for training other LLMs.

In this paper, we investigate and verify that instruction-tuning reduces the output diversity of various open-source and open-weight LLMs when applied to an open-ended narrative generation task.
Specifically, we look at OLMo 0724 7B \citep{groeneveld-etal-2024} and OLMo 2 7B 1124 \citep{olmo-2025} more in-depth and show the ``diversity gap'' can be attributed to different stages of preference tuning.
Motivated by our findings, we propose conformative decoding, a strategy designed to reintroduce diversity by attenuating the instruct model using its more diverse base model, thus allowing the usefulness of instruction-tuned LLMs while enabling more diverse outputs.

The key contributions are:
\begin{enumerate}
    \item Through narrative generation tasks, we demonstrate the existence of a diversity gap due to instruction-tuning in five LLMs.
    \item We show that DPO \citep{rafailov-etal-2023-direct} is primarily responsible for the diversity gap, as it considerably reduces the output diversity across the generated narratives.
    \item Finally, we present conformative decoding, a method to reintroduce output diversity. Our results show significant improvements while maintaining quality.
\end{enumerate}

\section{Background}
Diversity in generative AI outputs is receiving more attention, especially on
evaluation of the diversity of their outputs \citep[e.g.,][]{pillutla-etal-2023,le-bronnec-etal-2024}. 
For LLMs, there are various methods and decoding strategies to make them into more diverse generators \citep[e.g.,][]{vijayakumar-etal-2018,holtzman-etal-2020,li-etal-2023}.
In this section, we discuss what diversity is, how to measure it, and review efforts to make LLMs more diverse.

\subsection{What is diversity?}
Diversity is a complex and dynamic concept. 
It has different meanings in different research areas shaped by the goals and challenges.
Generally, diversity represents the presence of differences or variety in some context along some dimension.
More strictly, diversity is a quality observed in a set of objects; a single entity cannot be considered diverse.

One view of diversity based on entropy is used in ecology to quantify biodiversity \citep{leinster-2021}. 
A population is diverse because it has a wide range of varieties.
Language is typically concerned with two kinds of diversity: lexical diversity and semantic diversity. 
The former is more about style or form, while the latter is about content.
Diversity plays an important role in creative domains \citep{jordanous-keller-2016}; for example, if a band is known for its diverse music, it means their songs are not confined to a single genre.
In information retrieval, diversity is studied specifically for recommender systems, which is necessary to satisfy users. 
However, users do not always desire the same kind of diversity; some may want diversity within genre, and others may want to explore beyond familiar genres \citep{robinson-brown-schedl-2020}.

When is a generative AI model diverse?
Usually, when a model produces a broad and diverse set of outputs, it is more useful in different kinds of situations.
A model that only generates images or texts in the same style or recommends the same songs again and again, is not very diverse or useful.
Another view is that diversity could be about representational diversity.
This line of research investigates how generative models depict minority groups \citep{shihadeh-ackerman-2023a,ackerman-brown-2024} or associate certain human characteristics with a certain demographic \citep{fraser-etal-2023,shihadeh-ackerman-2023b}.
While we highly appreciate the valuable work being done on representation diversity, this paper focuses on linguistic diversity.

\subsection{Measuring Quality and Diversity}\label{ssec:measuring-diversity}

Quality and diversity are two key criteria for assessing the creativity of outputs \citep{jordanous-keller-2016,ibarrola-grace-2024}, and usually, there is a trade-off between the two \citep{le-bronnec-etal-2024}.
Models not only need to generate good outputs, but they must also produce a wide variety to be meaningfully integrated into the creative workflow, for example, generating many samples from a single prompt for exploration.

Evaluating generative AI models focuses on the same two axes: quality and diversity. 
The metrics generally require embedding the generated artefacts in a latent space.
A common way to measure quality is to use the Fr\'echet distance using the latent space of another model. 
For example, the Fr\'echet Inception Distance (FID) for measuring image quality uses the latent space of the InceptionV3 model \citep{heusel-etal-2017}.
Another method designed for assessing quality and diversity in generative models, respectively, is Precision and Recall using distributions \citep{sajjadi-etal-2018}.
The Improved Precision and Recall \citep{kynkaanniemi-etal-2019} metric works by drawing a hypersphere at each data point with a radius up to its $k$-th nearest neighbour to estimate manifolds for the generated samples $P_g$ and human reference $P_r$.
Precision is the probability that a point in $P_g$ falls within the support of $P_r$, and vice versa for Recall.
\textsc{Mauve} \citep{pillutla-etal-2021} is a measure specifically designed for text, which also compares a generated and a ground truth distribution using KL divergence in a quantised embedding space. 
\citet{pimentel-etal-2023} and \citet{pillutla-etal-2023} have shown that \textsc{Mauve} is an effective way to evaluate text.
While \textsc{Mauve} is designed to capture both quality and diversity, it cannot disentangle them.
Low \textsc{Mauve} scores could both be due to the lack of variety (e.g. repetitiveness, limited vocabulary) or poor quality (i.e. badly constructed sentences, wide vocabulary).

Approaches focussing specifically on measuring diversity have received less emphasis than those that measure quality.
As mentioned, \textit{Recall} is most commonly used to assess diversity. 
However, having a human reference that is considered diverse is essential so that if we have good coverage of the generated samples, they must also be diverse.
This approach is taken by \citet{hamalainen-tavast-kunnari-2023} and \citet{le-bronnec-etal-2024} to conclude that instruction-tuning lowers output diversity.

The metrics mentioned earlier all have two significant limitations. 
First, they all require a large sample size to get reasonable results. 
Secondly, they require ground-truth data.
In reality, we often want to evaluate small sets of generated samples, and obtaining a ground truth is usually difficult.
Two measures designed for diversity that do not have these drawbacks are the \textit{Vendi Score} \citep{friedman-dieng-2023} and \textit{Truncated Entropy} \citep{ibarrola-lawton-grace-2024}.
Both metrics take an entropic view of diversity, but each takes a different approach.
The Vendi Score is defined (\autoref{eq:vendi-score}) as the exponential entropy of the eigenvalues $\lambda$ of similarity matrix $\mathbf{K}/N$ where $\mathbf{K} \in \mathbb{R}^{N\times N}$.
The similarity function $k: \mathcal{X} \times \mathcal{X} \to \mathbb{R}$ computes $\mathbf{K}$ requiring $k(x,x) = 1$ and $k(x,y) = k(y,x)$.
\begin{equation}\label{eq:vendi-score}
    \mathrm{VS}_k(\mathcal{X}) \triangleq \exp{\left(- \sum^N_{i=1} \lambda_i \log \lambda_i \right)}
\end{equation}

Truncated Entropy was originally introduced as Truncated Inception Entropy as it relies on InceptionV3 to measure image diversity. Following the naming convention of FID, it similarly assumes normally distributed data. 
The metric is based on the differential entropy of a multivariate normal distribution.
As \citet{ibarrola-lawton-grace-2024} note that empirical approximation $\smash{\hat\Sigma}$ of covariance matrix $\smash{\Sigma}$ is singular and its determinant zero, when the number of samples $N$ is less than the dimensions of the latent space, making it infeasible for computing diversity for small sets of artefacts.
Instead, they proposed to truncate differential entropy (\autoref{eq:truncated-entropy}) and consider only the set of $N$ largest eigenvalues of $\smash{\hat\Sigma}$ to make computation feasible, where $l$ denotes the latent space which embeds the artefacts.
Originally designed to measure image diversity, its extension to CLIP included an experiment into text diversity \citep{ibarrola-grace-2024}.
\begin{equation}\label{eq:truncated-entropy}
\mathrm{TE}_l(\mathcal{X}) \triangleq \frac{N}{2} \log(2 \pi e) + \frac{1}{2} \sum^N_{i=1} \log \lambda_i^{(l)}
\end{equation}

A benefit of the Vendi Score is that it allows different similarity kernel functions $k$ to assess the desired flavour of diversity. 
Moreover, it is bounded between 1 and $N$, making it more natural to interpret, while Truncated Entropy is not bounded and can be negative due to small eigenvalues.

The above metrics all require some embedding in a latent space.
Other metrics that measure quality and diversity and which do not use text embedding features are perplexity and Zipf's Law \citep{zipf-1949} for quality, and $n$-gram diversity \citep{li-etal-2016}. 
However, perplexity is not a good measure for text quality \citep{nadeem-etal-2020,pillutla-etal-2021}, and the Zipf Coefficient is more a measure of closeness to natural language than quality per se. 
While $n$-gram diversity is a standard diversity measure, it has some drawbacks. 
If the sequences are of different lengths, the diversity of a set can be overestimated if it contains longer sequences with many unique words and conversely underestimated for short sequences with many repeating occurrences \citep{friedman-dieng-2023}.

\subsection{Improving the Output Diversity of LLMs}\label{ssec:improving-diversity}
Efforts to improve the output diversity of LLMs typically focus on decoding strategies. 
Decoding is the process of sampling a token from the next-token probability distribution given by the LLM on input.
These strategies are generally divided into two flavours: sampling and search-based methods.

Temperature sampling is the easiest method to increase output diversity, where increasing the temperature hyperparameter flattens the sampling distribution, but at the cost of quality \citep{peeperkorn-etal-2024}.
Often, temperature sampling needs to be complemented with a truncation strategy such as top-$k$ \citep{fan-etal-2018} or nucleus sampling \citep{holtzman-etal-2020} to prevent or at least reduce the risk of producing degenerate outputs.
The latter strategy, nucleus sampling, is interesting as it produces good quality text that is on par with diversity in human texts \citep{holtzman-etal-2020}.
Other sampling strategies perform similarly, as measured using $n$-gram diversity, with temperature sampling scoring best \citep{meister-etal-2023}.
Sampling decoding strategies generally do not focus specifically on optimising output diversity.

Search-based methods use beam search (often combined with a sampling strategy) to track multiple candidates and score them based on probability over multiple steps.
The goal is to find good sequences with a high but ``hidden'' probability that might be missed when decoding using sampling only.
An effective search-based method that yields more diverse sequences is Diverse Beam Search \citep[DBS;][]{vijayakumar-etal-2018}. 
DBS considers a fixed number of groups of beams, each group lagging one timestep behind the previous group.
The objective function combines the joint probability of beams and the dissimilarity between groups of beams at each step.
This results in more diverse outputs compared to standard beam search. 
Contrastive Decoding \citep{li-etal-2023} is another search-based strategy that uses an objective where a larger model (expert) is penalised by a smaller model (amateur) to prevent the expert from following bad language patterns learned by the amateur.
Contrastive Decoding leads to more diverse outputs than other search-based methods but is less effective than nucleus sampling \citep{li-etal-2023}.
Search can increase output diversity but at a higher cost compared to sampling.

Another approach is to optimise the diversity of the output informed by multiple models using an ensemble of LLMs \citep{tekin-etal-2024}.
Depending on the input, this method selects a sub-ensemble out of a large ensemble of LLMs that is likely to yield a diverse output. 
Subsequently, the probability distributions of the sub-ensemble are combined but require an ensemble learner (another neural network) to yield diverse outputs.
Finally, there are prompt-based approaches to increase diversity in various domains \citep{chu-etal-2024,chen-etal-2025}, but these are model- (and task-) specific, while we focus on a model-agnostic approach.

\section{The Diversity Gap: Does Instruction-tuning Reduce LLM Output Diversity?}\label{sec:diversity-gap}
Previous work observed that instruction-tuning lead models to yield less diverse outputs \citep{hamalainen-tavast-kunnari-2023,le-bronnec-etal-2024,kirk-etal-2024,chu-etal-2024}.
In this section, we investigate the existence of a ``diversity gap'' for various small LLMs performing a narrative generation task.
We explore the extent of the diversity gap using various diversity metrics, the impact of different fine-tuning steps on diversity, and the impact of conversation templates.
These experiments motivate our decoding strategy presented in \autoref{sec:conformative-decoding}, which aims to reintroduce some lost diversity in instruct models.

\subsection{Experimental Setup}

\subsubsection{Narrative Generation Task and Data}
The task in this paper is based on the concept of a writing prompt, a short statement or question, which is designed to spark and inspire a writer's creativity. 
The r/WritingPrompts subreddit is a space designed to foster creative writing and productivity. 
People post a short premise of a story they want to see written. 
In the comments, other people supply stories in response. 
We put this task to the LLMs and evaluate their performance in diversity and quality.

However, since LLMs entered the scene, it has become increasingly difficult to find sources of data untouched by LLM-generated data. 
This is problematic since many quality and diversity metrics rely on a human reference.
Fortunately, the Writing Prompts dataset \citep{fan-etal-2018} was compiled using prompts and stories from the subreddit before the widespread adoption of LLMs, providing a solid ground truth for our experiments.
We use a subset of the Writing Prompts dataset since we need sufficient human stories for each writing prompt to assess diversity. Specifically, we selected all prompts from the training data that have at least 50 human responses. 
We manually filtered the prompts that were inappropriate or offensive (see Appendix \ref{app:ethics}) and the responses that start with commentary of some sort (e.g., ``This is my first post'', or ``Feedback welcome.'').
We kept the first 50 human responses for each writing prompt, giving a total of 53 prompts and 2650 human responses.
Since the Writing Prompts dataset is considered diverse \citep{fan-etal-2018}, we use our subset as ground truth in metrics when required.

\begin{figure}[t]
     \centering
     \begin{subfigure}[b]{0.495\textwidth}
         \centering
         \includegraphics[width=\textwidth]{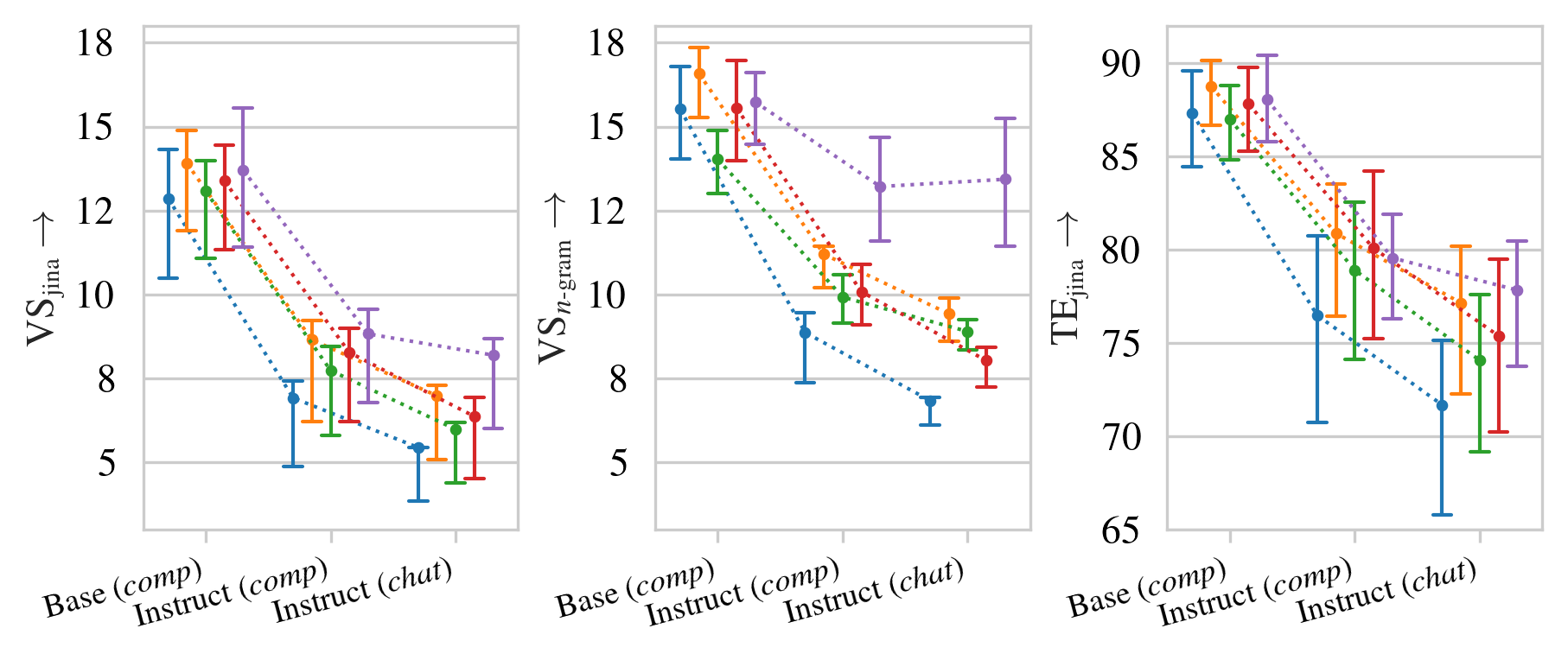}
         \caption{Metrics per-prompt}
         \label{fig:diversity-gap-a}
     \end{subfigure}
     \begin{subfigure}[b]{0.495\textwidth}
         \centering
         \includegraphics[width=\textwidth]{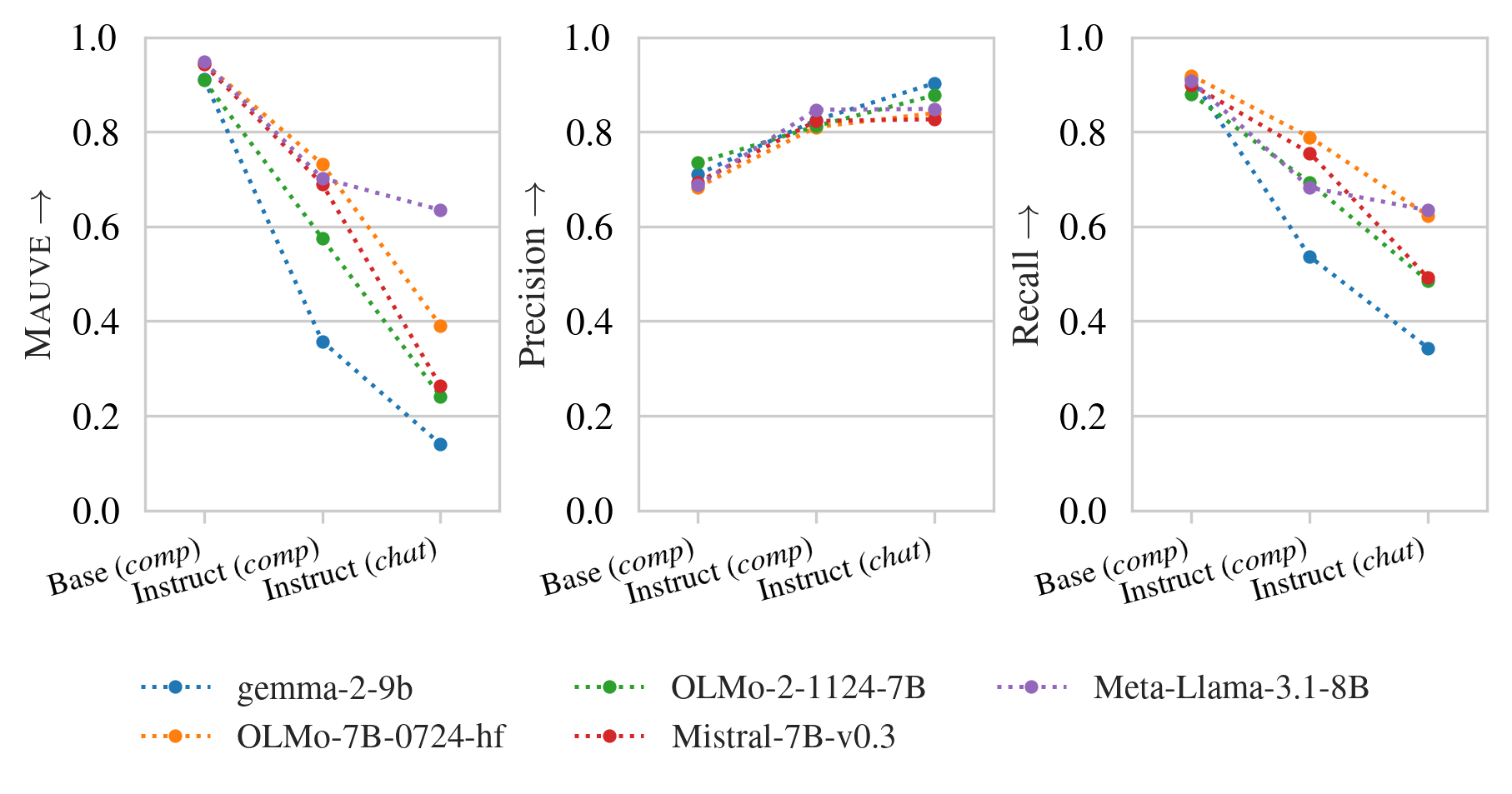}
         \caption{Metrics across-prompts}
         \label{fig:diversity-gap-b}
     \end{subfigure}
     \caption{We compare the base and instruct models for both prompts, \textit{comp} and \textit{chat}, on the narrative generation task. \autoref{fig:diversity-gap-a} shows a drop in diversity for all metrics for each model ($p < .001$). The chat template typically further reduces output diversity. In \autoref{fig:diversity-gap-b}, \textsc{Mauve} and Recall further show that instruction tuning reduces output diversity while precision increases, suggesting a trade-off between quality and diversity. Error bars represent the 0.5 percentile interval.}\label{fig:diversity-gap}
\end{figure}

\subsubsection{Prompting and Incipits}\label{sssec:prompting-and-incipits}
Base and instruct models require different prompts to perform a task, therefore, we have a text completion prompt and a conversation prompt (\autoref{fig:completion-prompt}). 
The base models are always prompted such that their natural response is to complete a story, denoted as (\textit{comp}). 
In the case of instruct models, we either use the completion prompt or prompt using the conversation template, denoted as (\textit{chat}). 
Allowing us to deepen our understanding of the difference in diversity between a completion and a conversation prompt for instruct models.
To this end, we must ensure consistent behaviour between the base and instruct models, i.e., they must generate outputs of the correct artefact class: stories.
Instruct models using the conversation prompt have no problem performing the task.
Generating stories using the completion prompt is much harder, especially for base models, which often output degenerate, repetitive text or items that are not a story.
To alleviate this unwanted behaviour, we use incipits of the human stories in the dataset to naturally guide the models into their completion ``umwelt'' (its model of the world). 
These incipits introduce a bit of diversity but ensure the models properly perform the task on both prompts.
The incipit size is 20 tokens.
For consistency, we use the tiktoken library with the \texttt{o200k\_base} encoding to truncate the human responses into incipits for all models.
Few-shot prompting \citep{brown-etal-2020} is beyond the scope of this paper, as it introduces much natural variation, making comparison difficult.
The completion and conversation naturally have some differences, but we aim for the prompts to be as simple and similar as possible.

\begin{prompt}[t]
\small
\begin{lstlisting}
Writing prompt: [WP]

The story is as follows: [incipit]
\end{lstlisting}
\caption{The exact prompt for text completion, \texttt{[WP]} and \texttt{[incipit]} are replaced with the writing prompt and the first 20 tokens of a human response, respectively. In the conversation prompt, we replace ``The story is as follows: '' with the instruction  ``Write a story'' and apply the model-specific chat template, leaving the rest, including the incipit, unchanged.}\label{fig:completion-prompt}
\end{prompt}

\subsubsection{Models}\label{sssec:models}
For practical reasons, we focus our experiments on smaller open-weights and open-source models. 
In particular, we look at Gemma 2 9B \citep{gemma-team-2024-gemma2}, Meta Llama 3.1 8B \citep{grattafiori-etal-2024-llama3}, Mistral 7B v0.3 \citep{jiang-etal-2023-mistral7b}, OLMo 7B 0724, and OLMo 2 1124 7B.
The OLMo families are fully open-source and provide intermediate checkpoints of each fine-tuning step, giving the opportunity to see how much diversity is lost in the Supervised Fine-tuning (SFT), Direct Preference Optimisation \citep[DPO;][]{rafailov-etal-2023-direct}, and Reinforcement Learning with Verifiable Rewards \citep[RLVR;][]{lambert-etal-2025-tulu3} steps.
As such, we run experiments to investigate the OLMo models in further detail. 
For all experiments, we use nucleus sampling ($p = .95$) with temperature $t = 1$.
For each writing prompt in the Writing Prompts subset and each configuration: Base (\textit{comp}), Instruct (\textit{comp}) and Instruct (\textit{chat}), we generate 50 samples using 50 incipits, up to a maximum of 500 tokens per story.

\subsubsection{Evaluation Metrics}\label{ssec:eval-metrics}
We evaluate the generated text using a range of quality and diversity metrics (lexical and semantic).
To measure lexical diversity, we use the Vendi Score (\autoref{eq:vendi-score}) with $n$-gram counts for $n \in \{1,2,3,4\}$ as features to compute the similarity matrix, denoted as \vsn.
For semantic diversity, we use the Vendi Score and Truncated Entropy (\autoref{eq:truncated-entropy}) applied to text embeddings (see below), denoted as \vsj and \te, respectively.
We use raw text embeddings for \te as normalised features yielded unintuitive results.

We use Improved Precision and Recall \citep{kynkaanniemi-etal-2019} using $k=3$ to compute Precision and Recall as quality and diversity metrics, respectively.
High precision suggests realistic and accurate generated texts, while high Recall suggests good coverage of the ground-truth reference, implying high diversity.
We evaluate the stories using \textsc{Mauve} with the default settings and follow best practices\footnote{See \url{https://github.com/krishnap25/Mauve?tab=readme-ov-file\#best-practices-for-Mauve}}. We use the embedding model mentioned below to compute the text features instead of the default GPT2-large, as better embedding models generally lead to better results. 
Note that we compute both \textsc{Mauve} and \textit{Improved Precision and Recall} across all prompts since these methods require many samples to give an accurate measurement. 
The Vendi Score and Truncated Entropy are applied on a per-prompt basis.

To obtain the text embedding features, we use the \texttt{jinaai/jina-embeddings-v3} model \citep{sturua-etal-2024}.
This embedding model is the highest scoring\footnote{At the time of writing, accessed on 13 January 2025 at \newline\url{https://huggingface.co/spaces/mteb/leaderboard}} model on Semantic Textual Similarity (STS) task and on average on the MTEB benchmark \citep{muennighoff-etal-2023} making this a suitable embedding model for our purpose. 

Finally, we use a one-tailed paired t-test to test statistical significance for the \vsj, \vsn, and \te results, with the hypothesis that the mean diversity of the base model is greater than that of the instruct model.
We compare the Base (\textit{comp}) against the Instruct (\textit{comp}) and Instruct (\textit{chat}).
Finally, to measure the effect of the chat template, we test the Instruct (\textit{comp}) against Instruct (\textit{chat}).

\subsection{Results for Diversity Gap Experiments}\label{ssec:div-gap-results}

\paragraph{Does the diversity gap exist?}
\autoref{fig:diversity-gap-a} shows that along the \vsj, \vsn, and \te, there is a significant drop in diversity for outputs generated by instruct models ($p<.001$ for all comparisons).
The diversity gap is also reflected in the \textit{Recall} scores with a clear drop. On the other hand, \textit{Precision} improves (\autoref{fig:diversity-gap-b}), suggesting that the generator is better at producing realistic outputs. 
This shows that the generated text fits well within the reference manifold.
However, the decline in \textit{Recall} shows that fewer reference texts fall within the generated manifold, indicating that instruction-tuning is a trade-off between quality and diversity. 
The \textsc{Mauve} scores for the base models are high, demonstrating a good balance of high quality and high output diversity.
This may be expected since the Writing Prompt dataset could easily be part of the training mix for any of these LLMs.
However, the other diversity scores for the instruct models are much lower (\autoref{fig:diversity-gap-a}), especially for the conversation prompt, suggesting that the drop for \textsc{Mauve} is likely to be due to the generator producing many similar outputs.
Overall, the results in \autoref{fig:diversity-gap} clearly show that instruction-tuning reduces the output diversity of LLMs.

\begin{figure}[t]
     \centering
     \begin{subfigure}[b]{0.495\textwidth}
         \centering
         \includegraphics[width=\textwidth]{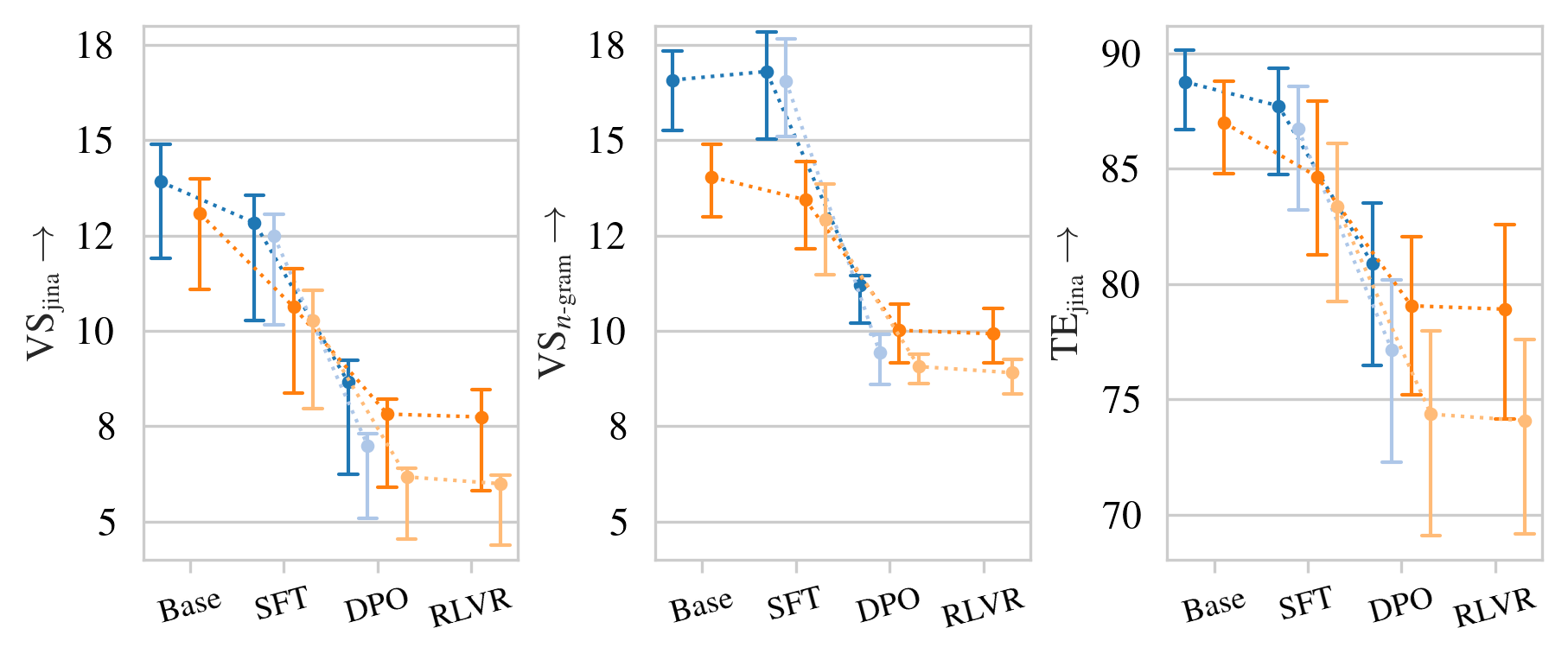}
         \caption{Metrics per-prompt}
         \label{fig:olmo-2-intermediate-gap-a}
     \end{subfigure}
     \begin{subfigure}[b]{0.495\textwidth}
         \centering
         \includegraphics[width=\textwidth]{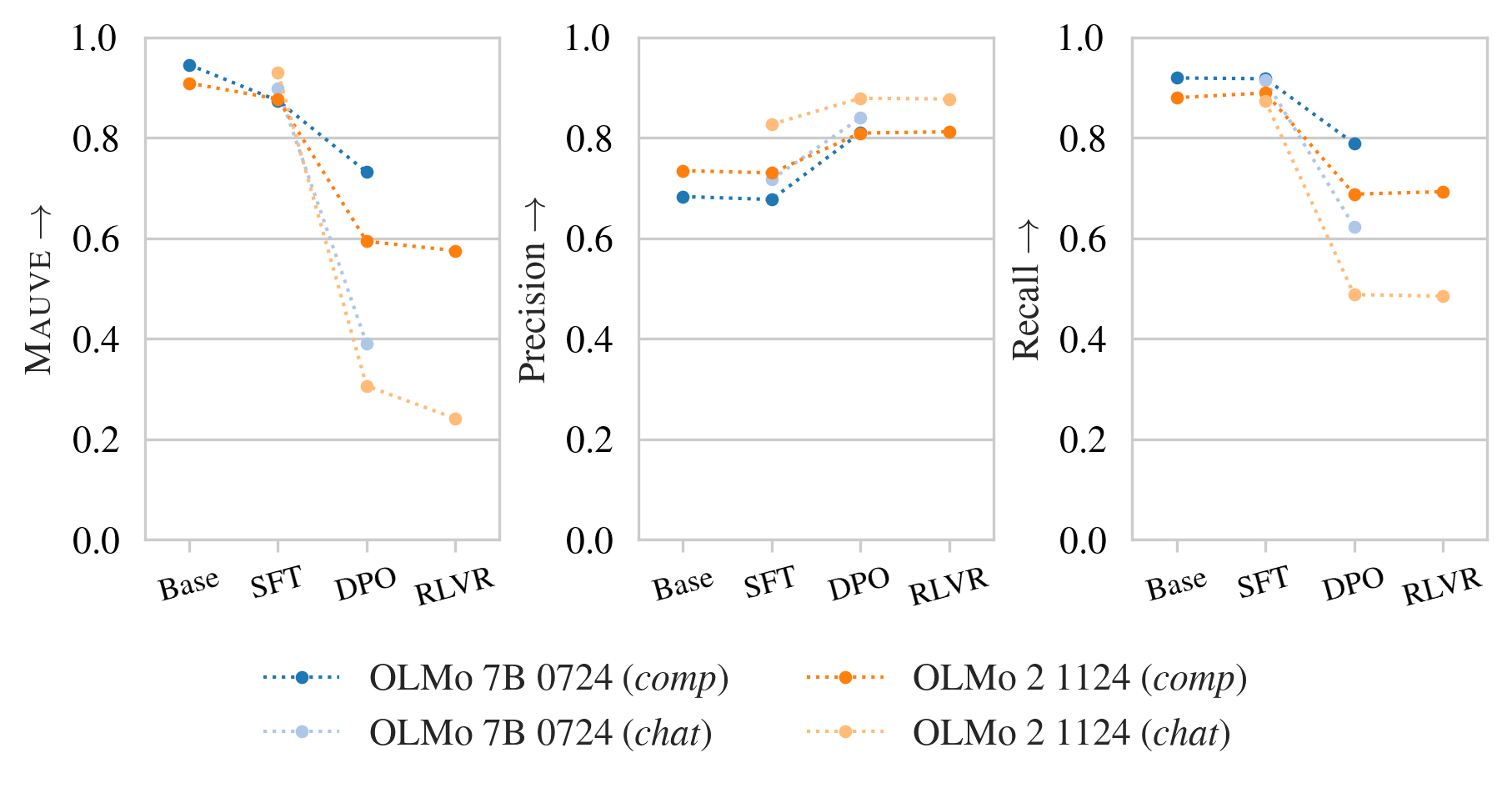}
         \caption{Metrics across-prompts}\label{fig:olmo-2-intermediate-gap-b}
     \end{subfigure}
      \caption{This figure presents experiments that investigate how diversity is lost with each fine-tuning step for OLMo 0724 7B and OLMo 2 7B 1124. \autoref{fig:olmo-2-intermediate-gap-a} shows an increasing decline in diversity with each step. In \autoref{fig:olmo-2-intermediate-gap-b}, there is a steep drop for the DPO models as measured by \textsc{Mauve}, and Recall suggests that reward-based fine-tuning has a particularly negative effect on output diversity. Error bars represent the 0.5 percentile interval.}\label{fig:olmo-2-intermediate-gap}
\end{figure}

\paragraph{What is the impact of each fine-tuning step?}
To further strengthen the findings above, the in-depth experiments for OLMo 0724 7B and OLMo 2 7B 1124 show that both SFT and DPO impact the output diversity of the model but that the new fine-tuning step, RLVR, has little effect, as shown in \autoref{fig:olmo-2-intermediate-gap}.
Recall and \textsc{Mauve} \autoref{fig:olmo-2-intermediate-gap} reveals that DPO has a much stronger effect than SFT, showing that reward modelling according to human preference has a particularly negative effect on LLM output diversity.
Moreover, the Precision and Recall results hint that generated texts constitute a much smaller manifold than the reference due to DPO; they suggest a direction for future investigation.

\paragraph{What is the impact of conversation templates?}
Prompting the instruct models using the conversation prompt further pushes the model into a narrow output space, as it yields lower scores on diversity and quality compared to the completion prompt ($p<.01$ for all comparisons, except \vsn for Llama 3.1 8B).
This is expected as the chat template is part of training, and using the completion prompt on instruct models is less restrictive.

\section{Conformative Decoding}\label{sec:conformative-decoding}
Motivated by the existence of the diversity gap, we present a sampling decoding strategy that aims to reintroduce some diversity in the outputs of instruct models.
The idea is that we can use base models to inform the instruct models and enable higher diversity while leveraging the usefulness and productivity of the instruction following models.

\subsection{Decoding Strategy}
Our sampling decoding strategy is a weighted sum to push the instruct model to conform to its more diverse base model (Equation \ref{eq:conf-decode}).
Hence, we call our strategy \textit{conformative decoding}.
It is essential that conformative decoding is used in combination with a truncation strategy, such as top-$k$ \citep{fan-etal-2018}, nucleus sampling \citep{holtzman-etal-2020}, or another truncation method, to mitigate the Softmax bottleneck \citep{yang-etal-2018}.
Conformative decoding is applied to the truncated probability distribution of the instruct model.
Initial explorations showed a greater increase of diversity applied before truncation, but this often results in low-quality outputs.
Applying the mixture after truncation is conceptually more elegant and increases diversity only when the instruct model provides room.
This prevents the degenerate outputs while reducing the magnitude of the conforming effect. 
We note, however, that truncation should not be too restrictive if aiming for diversity.


\begin{align}\label{eq:conf-decode}
    \mathrm{ConformativeDecoding}(x_t \mid x_{<t}) =& \nonumber \\
    \gamma\log p_{\theta}(x_t \mid x_{<t}) + (1 - \gamma)\log p_{\phi}&(x_t \mid x_{<t}) \nonumber \\
    \mathrm{\ for\ } x_t \in\ &\mathcal{V}_{\mathrm{valid}}
\end{align}

where $\theta$ are the parameters of the instruct model, and $\phi$ the parameters of its base model, $\gamma$ is a hyperparameter to control the mix of the two models, and $\mathcal{V}_{\mathrm{valid}}$ is the set of valid tokens according to the chosen truncation strategy. In this work, we use nucleus sampling with $p=.95$ where $\mathcal{V}_{\mathrm{valid}}$ is the minimal set that covers at least 95\% of the probability mass.
The log probability for $x_t \notin \mathcal{V}_{\mathrm{valid}}$ are set to $-\infty$, meaning they are not considered in the sampling process.
In practice, we apply the mixture using logits directly for efficiency, which is equivalent after further Softmax normalisation.

\begin{figure}[t]
     \centering
     \begin{subfigure}[b]{0.495\textwidth}
         \centering
         \includegraphics[width=\textwidth]{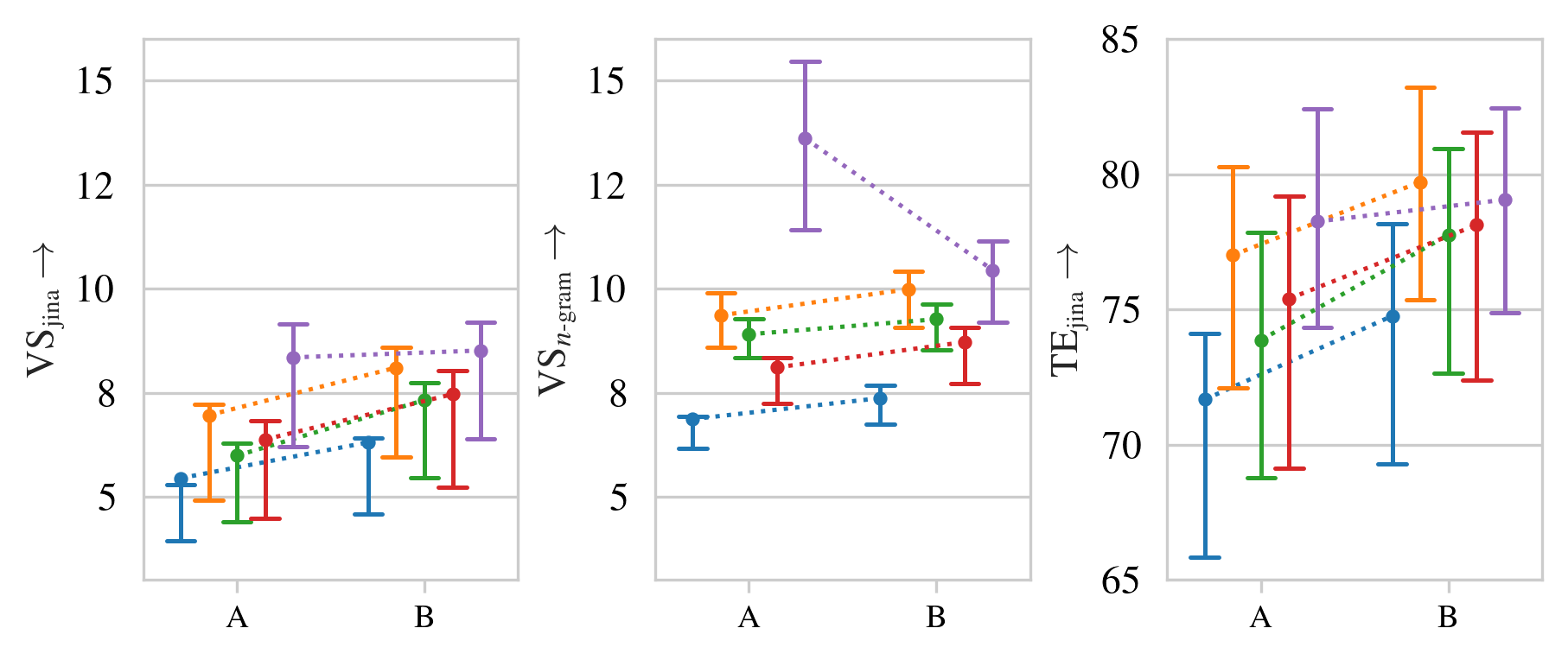}
         \caption{Metrics per-prompt}
         \label{fig:conformative-gap-a}
     \end{subfigure}
     \begin{subfigure}[b]{0.495\textwidth}
         \centering
         \includegraphics[width=\textwidth]{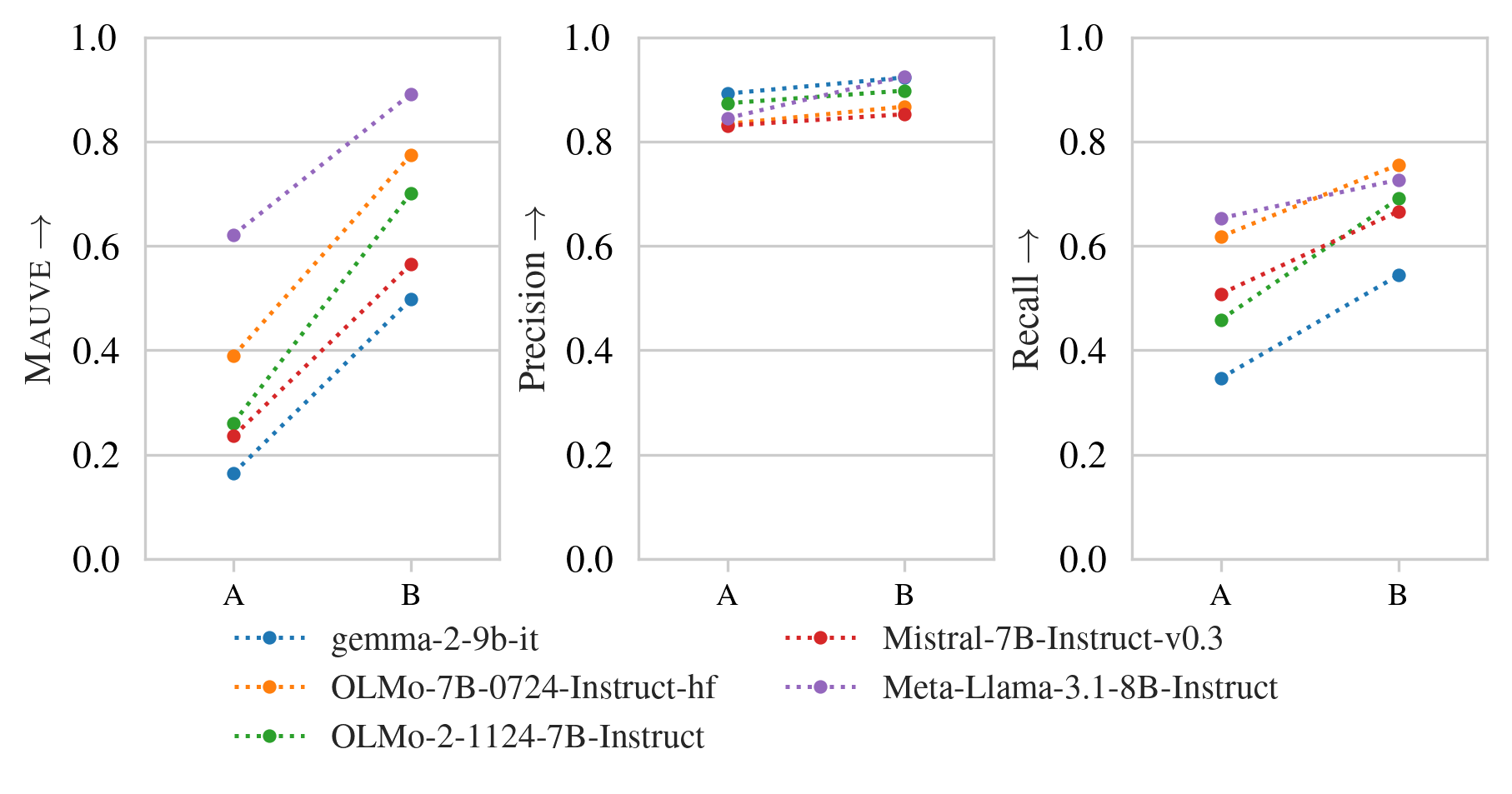}
         \caption{Metrics across-prompts}
         \label{fig:conformative-gap-b}
     \end{subfigure}
     \caption{This figure shows automatic evaluations for configuration A (nucleus sampling using $p=0.95$) and configuration B (nucleus sampling with conformative decoding using $p=0.95$ and $\gamma=0.5$). \autoref{fig:conformative-gap-a} shows that conformative decoding improves output diversity for all models ($p < .001$), except for Meta Llama 3.1 8B. In \autoref{fig:conformative-gap-b}, \textsc{Mauve} and Recall also show increased diversity, while quality is maintained as indicated by Precision, and even slightly improves. Error bars represent the 0.5 percentile interval.}\label{fig:conformative-gap}
\end{figure}

\subsection{Experiment Setup}

The experiment in this section investigates if conformative decoding improves the diversity of instruct LLMs.
They are conducted using an identical setup as described in \autoref{sec:diversity-gap}, except we investigate if our method improves diversity while maintaining quality compared to a baseline.
We use the conversation template for the prompt and evaluate the outputs using the same metrics, along with the incipit, to accurately align the base model's ``umwelt'' with the task.
We apply two configurations to the same narrative generation task.
Configuration A is the baseline using nucleus sampling ($p=.95$), and configuration B uses nucleus sampling with conformative decoding ($p=.95$, $\gamma=.5$).
We choose $\gamma=.5$ as this is a neutral value; higher $\gamma$ would increase diversity and vice versa, but we leave further hyperparameter tuning for future work.
Finally, to test significance we use a one-tailed paired t-test with the hypothesis that the baseline has a lower mean diversity than our method.

\subsection{Results for Conformative Decoding}\label{ssec:conf-dec-results}

The hypothesis for these experiments is that, by having the instruct model conform to the base model, we reintroduce diversity in the LLM's outputs.
The result in \autoref{fig:conformative-gap-a} shows that for \vsj, \vsn, and \te, except for Llama 3.1 8B, we see significant increases in diversity for all models ($p<.001$). 
Llama 3.1 8B even measures a significant decline on \vsn ($p<.001$), but interestingly, across-prompts, we see a recovery of diversity and even a slight increase in quality.

The across-prompt metrics all have a significant increase for each model.
\autoref{fig:conformative-gap-b} shows that our method improves diversity and maintains Precision.
Overall, the improvements are significant but modest; however, this is not unexpected since we truncate the next-token distribution according to the instruct model, which still influences the generation process.
A less restrictive truncation strategy naturally introduces more diversity at the cost of quality.

\section{Discussion}
The result of the diversity gap experiments provides additional evidence to the diversity gap between base and instruct models previously identified by \citet{hamalainen-tavast-kunnari-2023,kirk-etal-2024,le-bronnec-etal-2024,chu-etal-2024}. 
Our work explores this issue explicitly in the creative domain of narrative generation.
The automatic evaluation presented shows clearly that the diversity gap exists for multiple instruct LLMs. 
Instruction-tuning leads the model to focus on the conversation task and narrow its output space, which is further limited by the chat template.
Fine-tuning toward human preference, i.e. ``how'' the model responds, is a strong signal further limits output diversity.

Moreover, the loss of diversity suggests that models focus too heavily on the preference data. 
Plasticity, the observed effect that neural networks eventually lose their ability to learn from new data, is necessary for neural models to keep learning.
Yet, \citet{dohare-etal-2024} showed that is often not the case, implying that, at some point, fine-tuning is ineffective due to limited plasticity resulting from diversity loss.
This could have further implications, especially for RLVR, since our experiments \autoref{fig:olmo-2-intermediate-gap} show that DPO already significantly reduces diversity; it may be too late to adjust and learn more.
This as an interesting avenue for future work.

Conformative decoding shows improvements in both diversity and quality, which is promising for several applications that value diversity.
Yet, it proved less effective for Llama 3.1 8B as measured by \vsn, \vsj, and \te (\autoref{fig:conformative-gap-a}).
In \autoref{fig:diversity-gap-a}, Llama 3.1 8B shows the weakest drop in diversity; hence, it might simply be a better model to begin with.
\citet{zhou-etal-2023-lima} showed that Llama \citep{touvron-etal-2023-llama} could be improved by training on fewer but higher quality and highly diverse data to improve model performance. 
Similarly, this suggests that our method is primarily suitable for models that exhibit low diversity after preference tuning; its effectiveness could depend on the degree of the diversity gap.
In the across-prompts results, we still see increased diversity, suggesting the per-prompt evaluation is smoothed across prompts.
A follow-up study should investigate the relationship between the degree of the diversity gap and the improvements due to conformative decoding.

Finally, Conformative decoding relies on a truncation strategy to avoid degenerate outputs \citep{holtzman-etal-2020}.
Here, we use nucleus sampling as our baseline; it could be worthwhile to investigate other truncation strategies \citep[e.g.,][]{fan-etal-2018,meister-etal-2023,finlayson-etal-2024}.
Moreover, our method can easily be adapted to an objective function for beam search similar to contrastive decoding proposed by \citet{li-etal-2023}.

\section{Conclusion}
We investigate the diversity gap for a narrative generation task.
For our experiments, we evaluated diversity on a per-prompt basis using the Vendi Score (both lexical and semantic features) and Truncated Entropy, and we apply \textsc{Mauve}, Precision and Recall to evaluate quality and diversity across-prompts.
We show that instruction-tuned LLMs suffer a loss in diversity compared to their base models which is mostly due to DPO.
Based on this finding, we propose a decoding strategy to reintroduce diversity after instruction-tuning; conformative decoding typically reintroduces diversity while maintaining quality. 

\section*{Limitations}


Diversity is a tricky concept to capture for narratives, especially for longer sequences; it is difficult to assess by human participants \citep{shaib-etal-2025-standardizing}.
Narratives span an introduction, followed by actions and a conclusion \citep{sharples-pérezypérez-2022-story}. 
Therefore, assessing the diversity of a set of narratives requires extensive reading, causing fatigue among participants.
Truncating to an assessable narrative length would heavily impact the perceived diversity.
In future work, we aim to address this limitation and research effective and scalable methods for evaluating diversity in narrative generation using human participants.
Developing an automated approach to perform these kinds of evaluations is essential.

This study focuses solely on smaller open-source and open-weights models for practical reasons. 
Smaller models are less powerful and more likely to benefit from our decoding strategy, but also because running the base and instruct models simultaneously increases the computational overhead.
However, \citet{le-bronnec-etal-2024} showed that Llama 2 70B Chat also suffers a diversity loss, and future work should investigate the effects of our decoding algorithm for larger LLMs and in different domains that value diversity.

Finally, using incipits introduces some diversity, meaning we are not generating identical prompts for the 50 samples we draw for each writing prompt.
As mentioned in \ref{sssec:prompting-and-incipits}, this is a necessary limitation for a fair comparison with the base models to make them perform the narrative generation task consistently.
Omitting the incipit leads to comparing artefacts of different classes, which is not meaningful.


\bibliography{custom}

\appendix

\section{Computational Costs of Experiments}\label{app:computational-costs}
All experiments were run on up to 3 Nvidia A100 GPU 80GB.
For the diversity gap experiments (\autoref{ssec:div-gap-results}), depending on the model, each run required an estimated 7 to 18 hours of GPU compute.
The confirmation decoding experiments (\autoref{ssec:conf-dec-results}), depending on the model, each required an estimated 18 to 24 hours of GPU compute.

\section{Use of Scientific Artefacts}\label{app:artefacts}
The experiments in this paper use a subset of the Writing Prompt dataset by FAIR originally presented in \citet{fan-etal-2018} and is available under an MIT licence.

\section{Ethical Considerations and Risks}\label{app:ethics}
Narrative generation, which, more broadly, has been known to give rise to writings that can potentially be considered offensive. 
While selecting a subset of the Writing Prompt dataset for evaluation, we removed any offensive or inappropriate prompts and incipits, to mitigate risks for future research that involves human evaluation.

\end{document}